\pdfoutput=1

\documentclass[11pt]{article}

\usepackage{ACL2023}

\usepackage{times}
\usepackage{latexsym}

\usepackage[T1]{fontenc}

\usepackage[utf8]{inputenc}

\usepackage{microtype}

\usepackage{inconsolata}

\usepackage{graphicx}
\usepackage{amsmath}
\usepackage{amssymb}
\usepackage{booktabs}

\usepackage{algorithm}
\usepackage{amssymb}
\usepackage{adjustbox}
\usepackage{url}
\usepackage{colortbl}
\usepackage{multicol}

\usepackage{amsmath}
\usepackage{lipsum}  
\usepackage{adjustbox}
\usepackage{booktabs}
\usepackage{multicol}
\usepackage{multirow}

\usepackage{algorithm}
\usepackage{adjustbox}
\usepackage{xcolor}
\usepackage{mathtools}

\usepackage{enumitem}
\usepackage[noend]{algpseudocode}
\usepackage{graphicx}
\usepackage{subcaption}
\usepackage{bbm}

\algnewcommand\algorithmicforeach{{for each}}
\algdef{S}[FOR]{ForEach}[1]{\algorithmicforeach\ #1\ \algorithmicdo}

\usepackage{siunitx}

\newcommand\crule[3][black]{\textcolor{#1}{\rule{#2}{#3}}}
\newcommand{\progressbarWidth}{3cm}
\newcommand{\progressbarHeight}{0.25cm}
\newcommand{\progressbar}[2]{\crule[#2]
            {\progressbarWidth*#1/100}
            {\progressbarHeight}\crule[gray!30]
            {\progressbarWidth*(100-#1)/100}
            {\progressbarHeight}}

\title{Client-Customized Adaptation for Parameter-Efficient Federated Learning}

\author{Yeachan Kim$^{1}$\thanks{\ \ These authors contributed equally to this work.}, Junho Kim$^{1}$\footnotemark[1], Wing-Lam Mok$^{1}$, Jun-Hyung Park$^{2}$, SangKeun Lee$^{1,3}$ \\
$^{1}$Department of Artificial Intelligence, Korea University, Seoul, South Korea \\
$^{2}$BK21 FOUR R\&E Center for Artificial Intelligence, Korea University, Seoul, South Korea\\
$^{3}$Department of Computer Science and Engineering, Korea University, Seoul, South Korea \\
  \texttt{\{yeachan,monocrat,wlmokac,irish07,yalphy\}@korea.ac.kr}}

\begin{document}
\maketitle
\begin{abstract}
Despite the versatility of pre-trained language models (PLMs) across domains, their large memory footprints pose significant challenges in federated learning (FL), where the training model has to be distributed between a server and clients. One potential solution to bypass such constraints might be the use of parameter-efficient fine-tuning (PEFT) in the context of FL. However, we have observed that typical PEFT tends to severely suffer from heterogeneity among clients in FL scenarios, resulting in unstable and slow convergence. In this paper, we propose \textbf{C}lient-\textbf{C}ustomized \textbf{A}daptation (C2A), a novel hypernetwork-based FL framework that generates client-specific adapters by conditioning the client information.  With the effectiveness of the hypernetworks in generating customized weights through learning to adopt the different characteristics of inputs, C2A can maximize the utility of shared model parameters while minimizing the divergence caused by client heterogeneity. To verify the efficacy of C2A, we perform extensive evaluations on FL scenarios involving heterogeneity in label and language distributions. Comprehensive evaluation results clearly support the superiority of C2A in terms of both efficiency and effectiveness in FL scenarios\footnote{Our code is available at \url{https://github.com/yeachan-kr/c2a}}.

\end{abstract}

\section{Introduction}
The advent of large-scale pre-trained language models (PLMs) for natural language processing (NLP) has led to exceptional performance across a broad spectrum of domains. 
However, the high memory requirements for PLMs impede their applicability to resource-constrained environments. These challenges are particularly evident in federated learning (FL), where model weights are transmitted between the server and clients to preserve data privacy \cite{federated,federated2}.

While recent FL studies have expanded the application of PLMs in various tasks, such as text classification \cite{empericalfed, fedsentiment, pretrained-multi}, language modeling \cite{fedlang}, and question answering \cite{fedqa}, communicating the training model among clients requires huge computational resources and bandwidth, presenting a significant challenge in terms of practicality.

Parameter-efficient fine-tuning (PEFT) approach is thereby a promising strategy for reducing communication costs in FL. Through tuning only a small fraction of parameters, such as adapter-based tuning \cite{adapter, lora, compactor}, bias tuning \cite{zaken2022bitfit}, and prompt-tuning \cite{prompt}, PEFT approaches significantly enhance the memory efficiency in centralized scenarios. However, the feasibility of PEFT in decentralized scenarios has not been well explored.

\begin{figure}[t]
\centering
\includegraphics[width=\linewidth]{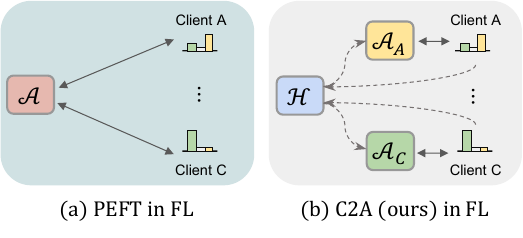}
\caption{Conceptual illustration of the existing PEFT modules ($\mathcal{A}$) and the client-customized adaptation ($\mathcal{H}$). The proposed method learns to generate the client-customized PEFT modules rather than fitting a single global module to all clients.
} 
\label{Figure1}
\end{figure}

Hence, we investigate the applicability of typical PEFT approaches in FL scenarios. Specifically, we measure the performance and \textit{client drifts} \cite{karimireddy2020scaffold,li2021model} of PEFT approaches in FL. 
Our discoveries are as follows: (1) typical PEFT approaches show large performance degradation in FL scenarios as the degree of non-IID increases; (2) these approaches usually suffer from large client drifts in non-IID scenarios, resulting in slow convergence and detrimental model performance. 
The above observations reveal that adopting PEFT in FL is not trivial, and posing the necessity to address large client drift.

To overcome the identified limitations, we propose a novel hypernetwork-based FL framework, \textbf{C}lient-\textbf{C}ustomized \textbf{A}daptation (C2A), that leverages the information of different data distributions on clients. Our key idea is to generate the adapter parameters tailored to each client via hypernetworks by taking the information of client data distribution, rather than naively fitting a single global adapter to all heterogeneous data distributions (Figure \ref{Figure1}). 
By learning to adopt the different data distributions to generate adapters for each client, C2A enables robust training for various non-IID conditions while sharing knowledge among clients. Moreover, in order to manage the large number of parameters associated with hypernetworks, we introduce factorized hypernetworks, thereby significantly reducing the number of parameters without sacrificing the performance. 

We carefully design the experimental setting to verify the efficacy of C2A on realistic FL scenarios, considering on both label and language heterogeneous. The experimental results show clearly that C2A can be robust to the heterogeneity of clients, thereby leading to the state-of-the-art results on diverse non-IID setups. In addition, our framework shows a significant enhancement in training efficiency across a range of downstream tasks. Finally, we demonstrate that our C2A successfully mitigates the large client drifts among local clients in non-IID scenarios. A summary of our main contributions is as follows:
\begin{itemize}
\item We investigate the effectiveness of PEFT among various FL scenarios. To the best of our knowledge, our work is one of the few researches for adapting PEFT in FL.

\item We propose \textbf{C}lient-\textbf{C}ustomized \textbf{A}daptation (C2A), a novel hypernetwork-based framework that strengthens the robustness of adapter concerning FL heterogeneity.

\item We demonstrate that C2A works quite well on various non-IID scenarios while preserving the benefits of efficiency in PEFT.

\end{itemize}

\section{PEFT in FL Scenario}

\subsection{Background of FL}
The goal of federated learning is to collaboratively train a single global model without sharing any private data between clients. To this end, FL proceeds through the communication of training models between clients and the server in a round-by-round manner. For each round, the server first distributes a single global model $\theta$ to a set of sampled clients, participating clients then perform local optimization on their own data. Upon the completion of the optimization, the server again aggregates all locally-trained models to update the global model. Formally, let the dataset of the ii-th client be $\mathcal{D}_i$, the above process for updating the global model can be formulated as follows:

\begin{equation}\label{eq:aggregation}
    \widetilde{\theta} = \sum_{i=1}^{K} \alpha_i \cdot \mathcal{L}(\mathcal{D}_i; \theta),
\end{equation}
where $\mathcal{L}(\mathcal{D}_i; \theta)$ is the function that returns the trained model based on the given dataset and the initial model, $K$ is the number of participating clients, and $\alpha_i$ is the contributing factor of the client $i$ to build a global model, which is typically determined by the dataset size of each client, i.e., $\alpha_i = \frac{|\mathcal{D}_i|}{\sum_{i}|\mathcal{D}_i|}$. While there are various aggregation methods, we focus on FedAvg due to its wide applicability in the FL community \cite{karimireddy2020scaffold,li2021model,luo2021no}.

However, utilizing cumbersome PLMs for the communication process of FL poses two challenges. Firstly, the function $\mathcal{L}(\cdot)$ requires high computing resources due to the large number of trainable parameters associated with PLMs. Secondly, in the aggregation step (i.e., weighted summation), significant network bandwidth is required to transmit and receive the models. Therefore, it is crucial to find an optimal solution that can mitigate these constraints, providing a more efficient and less resource-intensive mechanism for FL with PLMs.

\subsection{Impact of Heterogeneity on PEFT}\label{sect:motive_hetero}

\begin{figure}[t]
\centering 
\subfloat[Different level of heterogeneity]{%
  \includegraphics[width=1.0\linewidth]{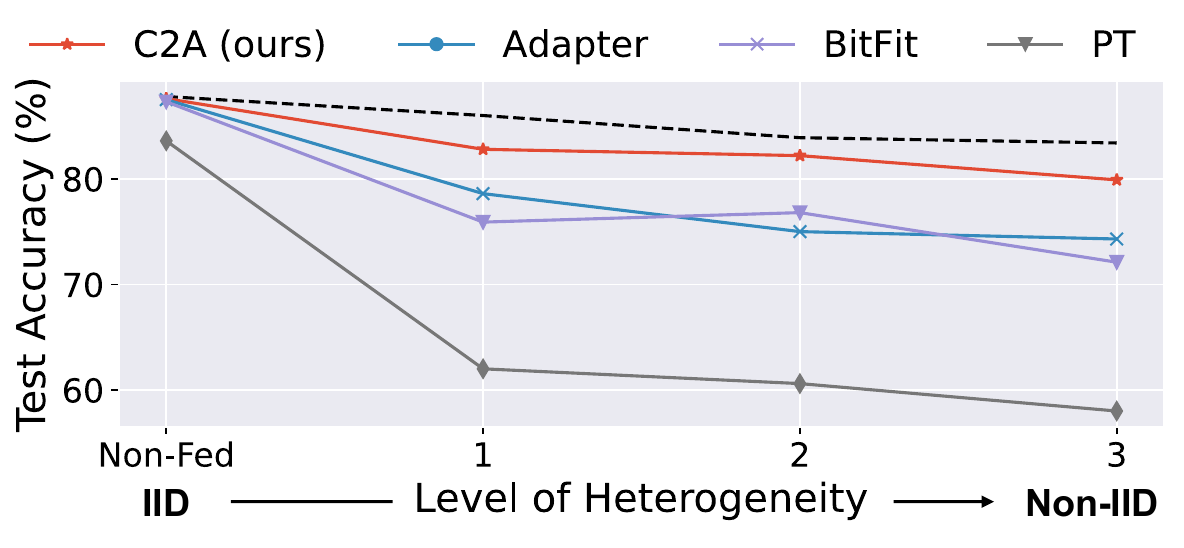}%
}

\subfloat[Drift from the global model]{%
  \includegraphics[width=1.0\linewidth]{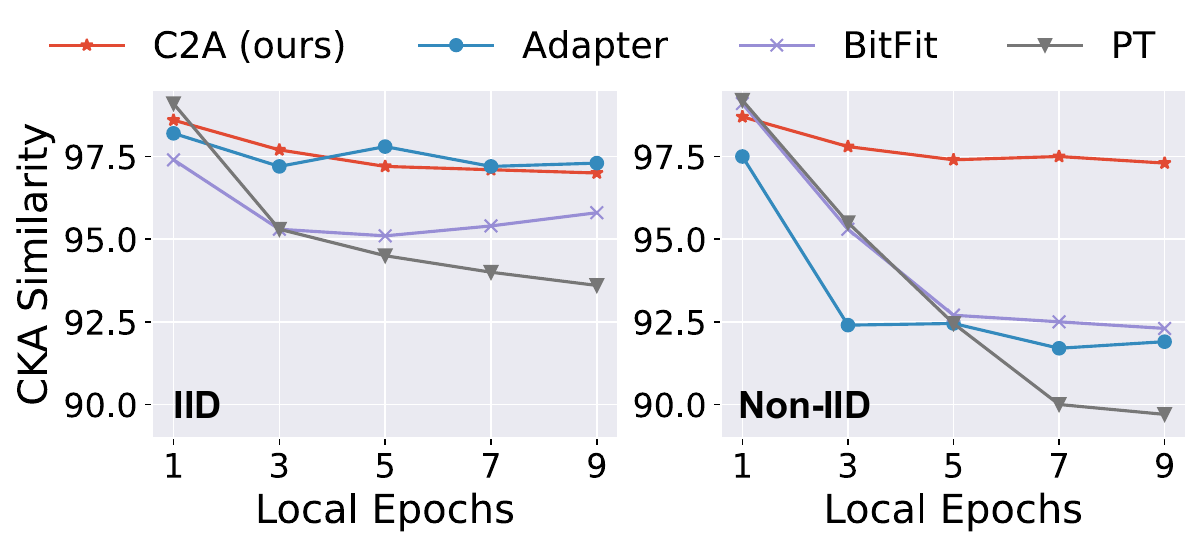}%
}
\caption{Sensitivity analysis of PEFT methods in federated context in terms of data heterogeneity and divergence from the global model. PT indicates prompt-tuning \cite{prompt}, and the dotted line indicates full fine-tuning method.}
\label{figure:motivation}
\end{figure}

To verify the applicability of PEFT in federated context, we conduct a preliminary investigation in which only small components (e.g., adapters, prompt embeddings, biases) are fine-tuned on local data and subsequently shared between clients. The experimental configuration comprises 100 clients engaged in the task of multilingual news classification\footnote{Further details for the experiment in Section \ref{subsection:setup}.} \cite{xglue}. 

We first examine the robustness of PEFT on heterogeneous data distribution between clients, which is common in real-world scenarios. We report the test accuracy of the global model with respect to the increasing heterogeneity\footnote{The increasing heterogeneity implies the corresponding degree of skewness deterioration towards certain classes}. The overall results are depicted in Figure \ref{figure:motivation}(a). In the non-federated scenario (i.e., IID), the existing PEFT methods manage to achieve strong performances comparable to that of the full fine-tuning. However, as the level of heterogeneity increases, the performances of the PEFT methods significantly lag behind that of the full fine-tuning. This verifies that PEFT methods exhibit greater susceptibility to heterogeneity than full fine-tuning. 



To gain a deeper understanding of the susceptibility, we further analyze the local optimization of the PEFT methods. Specifically, we measure the CKA similarity \cite{CKA} of the logits between the training model and the global model on the IID and non-IID setups. Figure \ref{figure:motivation}(b) shows the results. Comparing between IID and non-IID setups, all PEFT methods noticeably deviate from the global model on non-IID. This indicates that the model gradually converges to the client optima while drifting apart from the global model's optima, which are believed to be more generalized \cite{li2021model}. This observation aligns with prior results \cite{luo2021no}, and we suspect that such deviation attributes the slow and unstable convergence.





\section{C2A: Client-Customized Adaptation}

\begin{figure*}[ht]
\centering
\includegraphics[width=1\textwidth]{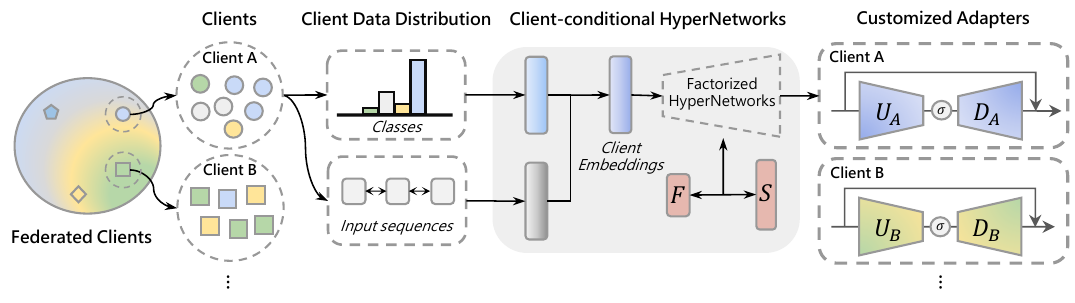}
\caption{Overview of the proposed framework, denoted as \textbf{C}lient-\textbf{C}ustomized \textbf{A}daptation (C2A). To perform customized adaptation, C2A takes into account the client information as a form of label and context. Based on the client embeddings, the factorized hypernetworks generate adapters that are specialized for each client.}
\label{Figure3}
\end{figure*}


In this section, we elaborate on the proposed framework in detail. The core strategy is to generate customized PEFT modules tailored to each client to mitigate the negative impact of heterogeneity among clients. To achieve this, we first derive latent vectors to represent the data distribution of each client (Section \ref{method:user}). The resulting embeddings are then conditioned on the hypernetworks so as to generate parameters of the PEFT modules tailored to each client (Section \ref{method:hyper}). Regarding on the large number of parameters induced from hypernetworks, we effectively factorize the weights of the hypernetworks (Section \ref{method:factorization}).



\subsection{Adapter Architecture}\label{method:adapter}
We start with defining the structure of the PEFT modules to be generated. While lots of different modules have been proposed, we focus on Adapter \cite{adapter}, given its versatility across domains, such as vision-and-image \cite{sung2022vl} and audio \cite{hou2021exploiting}, as well as its demonstrated efficacy in performing given tasks. The adapter consists of down- and up-projection functions that are interleaved between self-attention layers and feed-forward layers within every block of the PLMs. The adapting process can be formulated as:
\begin{equation}
    \mathcal{A}^l(x) = \textbf{U}^{l}\text{GeLU}(\textbf{D}^{l}x) + x
\end{equation}
where $\textbf{D}^{l} \in \mathbbm{R}^{r \times d}$ and $\textbf{U}^{l} \in \mathbbm{R}^{d \times r}$ are the weights for the down- and up-projection in the $l$-th layer of PLMs, respectively, $d$ is the hidden dimension of PLMs, and $r$ is the bottleneck dimension.



\subsection{Construction of Client Embeddings}\label{method:user}

To represent the characteristics of the clients, we consider two different types of information: 1) label embeddings and 2) context embeddings.


\paragraph{Label Embeddings} 

The label embedding plays a role in conveying the explicit information of class distribution on each client. Since mini-batches are generally sampled by uniform distribution, the label distributions on mini-batches can sufficiently represent the data distributions of clients. Thus we construct label embeddings from the label distributions of the mini-batches. Let the mini-batches of the client $i$ be $\mathcal{B} \subset \mathcal{D}_i$, the label embeddings can be derived as follows:
\begin{equation}
    L(\mathcal{B}) = \textbf{W}_{L}\text{avg}([y_1;...;y_{|B|}]) + b_{L},
\end{equation}
where $y_i$ is a one-hot label vector for the instance $x_i$, $[\text{ };\text{ }]$ denotes the concatenating function, $\text{avg}(\cdot)$ denotes the average pooling within mini-batches, $\textbf{W}_{L} \in \mathbbm{R}^{C \times t}$ and $b_{L} \in \mathbbm{R}^{t}$ are the linear transformation weights and biases for the number of classes $C$ and $t$ is the dimensionality of input embeddings. It is important to note that, since the labels for test data are not accessible, we opt for a uniform distribution for the inference phase to generate adapters that are not biased toward dominant classes.

\paragraph{Context Embeddings} 
Considering the contextual information in data can also provide an enhanced understanding of each client by taking a more comprehensive viewpoint (e.g., languages, text styles). Specifically, the contextual information is extracted from every layer to generate layer-specialized adapters. Inspired by the sentence embeddings \cite{li2020sentence}, context embeddings are extracted by averaging word vectors over the lengths with $\ell_2$ normalization. 
Let the resulting vectors of the sample $x_j$ from the $l$-th layer of PLMs be $f^{l}(x_j)$, the context embeddings of the $l$-th layer are derived as follows:
\begin{equation}
    F^{l}(\mathcal{B}) =  \textbf{W}_{F}\text{max}([f^{l}(x_1); ...; f^{l}(x_{|\mathcal{B}|})]) + b_{F}, 
\end{equation}
where $\text{max}(\cdot)$ denotes the max-pooling across the batch, and $\textbf{W}_{F} \in \mathbbm{R}^{d \times t}$ and $b_{F} \in \mathbbm{R}^{t}$ are the linear transformation weights and biases, respectively. 


\paragraph{Client Embeddings} The comprehensive client embeddings $\mathcal{I}^{l}_{\mathcal{B}}$ are constructed by summing up two types of embeddings. Additionally, we add layer-index embeddings into the client embeddings of each layer, further encouraging the generator to encode more diverse layer-wise information \cite{van2019does,de2020s}.


\subsection{Client-conditional HyperNetworks}\label{method:hyper}

Based on the client embeddings, we tailor adapters to each heterogeneous client. Drawing inspiration from the concept of hypernetworks \cite{hypernetwork} that generates parameters based on given input embeddings, we introduce the \textit{client}-conditional hypernetworks, which generate adapter parameters by taking the client embeddings $\mathcal{I}^{l}_{\mathcal{B}}$ as inputs. Formally, the parameters of the adapters (i.e., $U^{l}$, $D^{l}$) are generated by following the function of hypernetworks:
\begin{equation}\label{eq:no_facto}
    (\textbf{U}_{\mathcal{B}}^{l}, \textbf{D}_{\mathcal{B}}^{l}) := h(\mathcal{I}_{\mathcal{B}}) = (\textbf{W}_U, \textbf{W}_D)\mathcal{I}^{l}_{\mathcal{B}},
\end{equation}
where $\mathcal{I}$ is the input embeddings with dimensionality $t$, $W^{l}_{D} \in \mathbbm{R}^{(r \times d) \times t}, W^{l}_U \in \mathbbm{R}^{(d \times r) \times t}$ are the weights for the hypernetworks. Note that the hypernetworks are shared between different layers with the layer-specific information that are encoded to the input embeddings.

\subsection{Factorization of HyperNetworks}\label{method:factorization}

While customized adapters can be generated from the aforementioned hypernetworks, hypernetworks typically comprise a relatively large number of parameters. We thus factorize the proposed hypernetworks into two smaller weights. Moreover, the resultant matrices from the factorized components are $\ell_2$ normalized, such that the generated parameters are not biased towards any of the local majority classes in the client's data distribution \cite{zhong2021improving}. Formally, the up-projection weights in Eq.~\eqref{eq:no_facto} are reconstructed by two factorized components as follows:
\begin{equation}
    \textbf{U}_B^{l} = \textbf{W}_U\mathcal{I}_{B} = \sigma(\textbf{F}_U \textbf{S}_U)\mathcal{I}_{B}
\end{equation} 
where $\textbf{F}_U \in \mathbbm{R}^{d \times s}$ and $\textbf{S}_U \in \mathbbm{R}^{s \times (r \times t)}$ indicate the factorized components from $W_U$ with latent factor $s$, $\sigma(\cdot)$ denotes the Frobenius normalization.

For factorization, the latent factor $s$ plays a crucial role in determining the complexity and expressivity of the resulting adapters. To allow for a larger dimensionality of latent factors, the two projection weights are tied similarly as if the tied auto-encoder \cite{alain2014regularized}, i.e., $D^{l}_{B} = {U^{l}_{B}}^{\top}$. This strategy enables to halve the memory requirements without compromising the task accuracy.

\subsection{Aggregation Phase for C2A}

Upon the completion of the training phase on each client data, the respective trained models are transmitted back to the centralized server to update the global model (Eq.~\eqref{eq:aggregation}). Considering that the training models for C2A are hypernetworks, each client sends the parameters associated with the hypernetworks and the layer-index embeddings to the server in order to update the global hypernetworks.

\section{Evaluation}
In this section, we evaluate the efficacy of our C2A on two realistic FL scenarios: 1) heterogeneity in label distributions, and 2) heterogeneity in both label and language distributions.

\subsection{Datasets}
To simulate the two challenging scenarios, we mainly consider two text classification datasets, 20Newsgroup \cite{20newsgroup} and XGLUE-NC \cite{xglue}, which have recently served as benchmarks for evaluating FL for NLP \cite{fednlp,pretrained-multi}.

\paragraph{20Newsgroup } The dataset comprises 18,000 news posts that pertain to 20 distinct topics. Given its larger categorical space (i.e., 20 labels) than the typical sentiment analysis datasets, it is favored to the verification for the important factor of the label distribution heterogeneity scenarios.


\paragraph{XGLUE-NC } The dataset includes 10,000 posts written in multiple languages that pertain to 10 news categories. This diversity in languages adds an extra layer of complexity to the FL. The dataset comprises five languages: English, Spanish, French, German, and Russian. Furthermore, due to the varying categorical distribution between languages (e.g., the English dataset is skewed towards \texttt{Sports}, while the French dataset is skewed toward \texttt{News}), the distribution shifts among clients are naturally introduced to the dataset.

\subsection{Non-IID Client Partitioning} 
Building upon the two datasets, we adopt two non-IID partitioning strategies to inject heterogeneity into the label and language distributions.


\paragraph{Label Distribution. } Following the benchmark setup \cite{fednlp}, we apply Dirichlet distribution $\text{Dir}(\beta)$ to the datasets in reorganizing the data into the non-IID label distribution circumstance.
The value $\beta$ controls the degree of non-IID, the smaller the $\beta$, the more likely the clients in holding examples from only one class. Thus, we eventually construct a FL dataset respecting the label heterogeneity scenarios.

\paragraph{Language Distribution. } 
Following the language setup in \cite{pretrained-multi}, we randomly divide clients into five distinct groups, with each group being exclusively dedicated to a specific language. Subsequently, we split the dataset of each language in the same manner with the strategy of non-IID label distribution, which is more challenging and not even being explored in previous works.


\begin{table*}[h]
\centering
\caption{Evaluation results of test accuracy (\%) on 20Newsgroup. The best and second best results are highlighted in \textbf{boldface} and \underline{underlined}, respectively.}
\begin{tabular}{@{}lccccc@{}}
\toprule
\multirow{2}{*}{Methods} & \multirow{2}{*}{Params (\%)} & \multirow{2}{*}{Non-Fed} & \multicolumn{3}{c}{Federated scenario} \\ \cmidrule(l){4-6} 
                         &                            &                          & $\beta=5.0$    & $\beta=1.0$    & $\beta=0.1$    \\ \midrule
Full Fine-tuning & 100\% & 85.8  & 77.6  & 77.2  & 66.8      \\ \midrule
Adapter \cite{adapter} & 0.455\% & 84.0  & 69.1  & 65.5  & 56.1  \\ 
LoRA \cite{lora} & 0.111\% & \underline{84.3}  & \underline{69.5}  & \underline{67.7}  & \underline{56.6}     \\ 
Compacter \cite{karimi2021compacter} & 0.043\% & 83.2  & 65.9 & 62.8  & 50.1      \\
Prompt-tuning \cite{prompt} & 0.024\% & 74.2 & 51.6 & 46.4 & 28.2 \\
BitFit \cite{zaken2022bitfit} & 0.078\% & 82.8  & 67.1  & 66.5  & 55.1      \\ 
AdaMix \cite{adamix} & 0.559\% & \textbf{84.7} & 68.7 & 65.3 & 54.5 \\
\cellcolor[gray]{.9}C2A (ours.)                    & \cellcolor[gray]{.9}0.097\%              & \cellcolor[gray]{.9}83.9            & \cellcolor[gray]{.9}\textbf{71.6}   & \cellcolor[gray]{.9}\textbf{70.4}       & \cellcolor[gray]{.9}\textbf{61.0}      \\  \bottomrule
\end{tabular}
\label{exp:20news}
\end{table*}

\begin{table*}[h]
\centering
\caption{Evaluation results of test accuracy (\%) on XGLUE-NC. The best and second best results are highlighted in \textbf{boldface} and \underline{underlined}, respectively.}
\begin{tabular}{@{}lccccc@{}}
\toprule
\multirow{2}{*}{Methods} & \multirow{2}{*}{Params (\%)} & \multirow{2}{*}{Non-Fed} & \multicolumn{3}{c}{Federated scenario} \\ \cmidrule(l){4-6} 
                         &                            &                          & $\beta=5.0$    & $\beta=2.0$    & $\beta=0.5$    \\ \midrule
Full Fine-tuning         & 100\%                                & 87.6  & 84.5   & 83.7  & 80.7      \\ \midrule
Adapter \cite{adapter}  & 0.225\%                  & 87.5  & 78.6   & 75.0  & 74.3      \\
LoRA \cite{lora}                    & 0.055\%                   & \textbf{87.8}  & \underline{80.4}   & 78.4  & 74.6      \\
Compacter \cite{karimi2021compacter} & 0.021\%                  & 87.3  & 75.9   & 73.4  & 71.0      \\
Prompt-tuning \cite{prefix}   & 0.017\%                         & 85.6  & 61.2   & 60.6  & 58.0     \\
BitFit \cite{zaken2022bitfit}   & 0.038\%                       & 87.3  & 78.4   & 76.8  & 72.1      \\
AdaMix \cite{adamix} & 0.277\%                                  & \underline{87.6}  & 79.6   & \underline{79.1}  & \underline{76.6}   \\
\cellcolor[gray]{.9}C2A (ours.) & \cellcolor[gray]{.9}0.049\%  & \cellcolor[gray]{.9}87.4  & \cellcolor[gray]{.9}\textbf{82.8}   & \cellcolor[gray]{.9}\textbf{82.2}  & \cellcolor[gray]{.9}\textbf{80.2}      \\     \bottomrule
\end{tabular}
\label{exp:nc}
\end{table*}




\subsection{Federated Learning Setup}\label{subsection:setup}
\paragraph{Baselines and Implementations}
Following the previous work \cite{fednlp}, we use the uncased version of DistilBERT \footnote{In multi-lingual FL scenarios, we adopt the multi-lingual version of DistilBERT with 134M parameters} \cite{distilbert} with 66M parameters. We compare C2A with six strong baselines, which include Adapter \cite{adapter}, LoRA \cite{lora}, Compacter \cite{compactor}, Prompt-tuning \cite{prompt}, BitFit \cite{zaken2022bitfit}, and AdaMix \cite{adamix}, to encompass a broad range of PEFT methods. These modules are optimized by AdamW \cite{loshchilov2018decoupled} with the searched learning rate ranging from \{2e-4, 3e-4, 4e-4, 5e-4\}. 

\paragraph{Local Optimization and Aggregation}
We assign 100 clients for each dataset and randomly selected 25\% of the clients to join the local optimization in each round. During the local optimization, we use a batch size of 16 and 64 for 20Newsgroup and XGLUE-NC, respectively. Each client performs a single local epoch, and the server aggregates the locally-trained model based on FedAvg \cite{federated2}.


 

\subsection{Main Results}
To thoroughly evaluate each baseline on various FL setups, we start from a non-federated setup and progressively increase the level of heterogeneity by manipulating $\beta$. The results are shown in Table \ref{exp:20news} (20Newsgroup) and Table \ref{exp:nc} (XGLUE-NC).

The proposed method, C2A, achieves the state-of-the-art performance for almost all setups. 
Specifically, despite that AdaMix uses multiple adapters for ensemble, our model improves the respective performance by 3\% on both datasets. 
It is also noteworthy that while most PEFT approaches manage to achieve fair performance in non-FL scenarios, their performances significantly decrease as the degree of heterogeneity increases. In contrast, our C2A shows only marginal performance degradation even for high degree non-IID settings. Moreover, in the multilingual setting, C2A achieves a comparable performance to full fine-tuning. These results indicate that C2A is more resilient to heterogeneity in decentralized scenarios.


\section{Further Analysis on C2A}
In order to gain a deeper understanding of the benefits of C2A, we perform a series of analytical experiments utilizing XGLUE-NC with a value of $\beta = 0.5$, which represents the most challenging setup within our experimentation.

\subsection{Ablation Studies}
We conduct ablation studies to explore the contributions brought by each component of C2A. Specifically, we focus on the effect of client embeddings, which are composed of label embedding (LE), context embedding (CE), and factorization. Detailed results are presented in Table \ref{exp:ablation}. 

\paragraph{Client Embedding. } We observe that omitting either of the embeddings does hurt the model performance. Notably, comparing "w/o LE" to "w/o CE", ablating context embedding leads to more significant performance degradation. We suspect this is because that context embedding can provide more discriminating information of each client through implicit representations, such as language types, and text styles. Moreover, removing all the embeddings shows the worst performance, which demonstrates that our C2A with the client embeddings can generate more suitable adapters for each client.


\begin{table}[t]
\caption{Ablation studies for C2A. LE and CE represent the label and the context embedding, respectively.}
\adjustbox{width=\linewidth}{
\begin{tabular}{@{}lcc@{}}
\toprule
Methods            & Params (\%) & Accuracy (\%) \\ \midrule
\cellcolor[gray]{.9}C2A(ours)      & \cellcolor[gray]{.9}0.049\%        & \cellcolor[gray]{.9}\textbf{80.2}                \\ \midrule
\textit{Client embedding} \\
\ \ w/o LE      & 0.049\%           &  78.4      \\
\ \ w/o CE      & 0.049\%           &  78.0              \\
\ \ w/o LE,CE   & 0.049\%           & 77.3              \\ \midrule
\textit{Factorization} \\
\ \ w/o Factorization      & 0.106\%          & 79.8   \\
\ \ w/o Normalization      & 0.049\%          & 78.8   \\   
\bottomrule
\end{tabular}
}
\label{exp:ablation}
\end{table}

\paragraph{Factorization. } To examine the impact of factorization, we first compare it with the C2A results neglecting factorization. Despite using only half the parameters, our model achieves comparable performance as the model without factorization. In addition, we observe that omitting normalization significantly hurts performance. The results demonstrate that our normalization alleviates the performance drop by factorization.


\subsection{Local Epochs vs. Communication Rounds}

\begin{figure}[t]
\centering 
\includegraphics[width=1.0\linewidth]{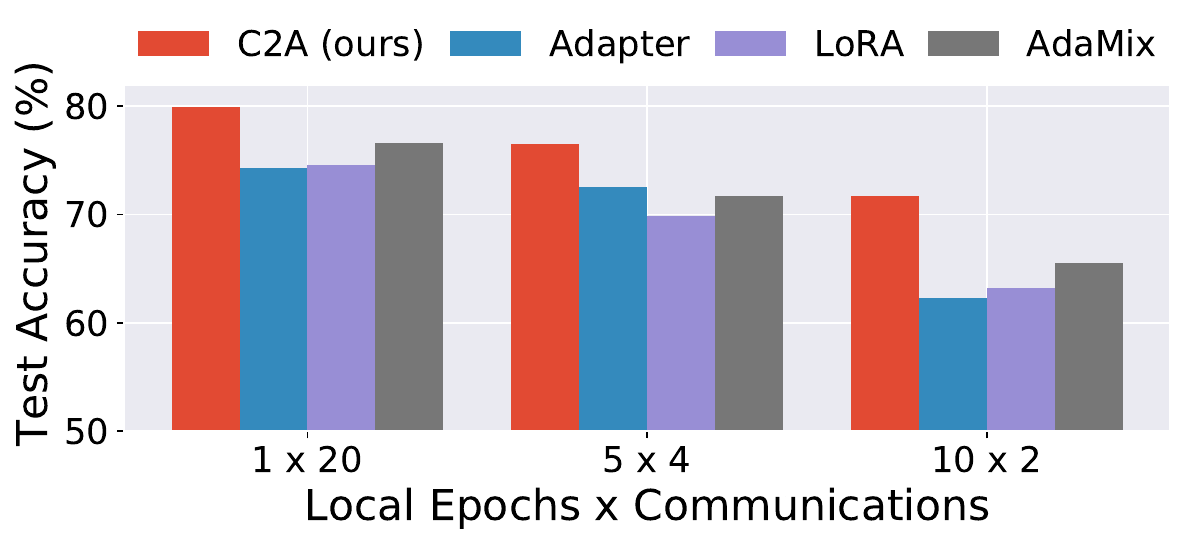}%
\caption{Evaluation results of the test accuracy with different numbers of local epochs.}
\label{analysis:num_epochs}
\end{figure}

One of the crucial aspects in FL is communication efficiency. A simple way to achieve such efficiency is to reduce communication rounds while increasing local epochs. However, the increased local updates can result in greater susceptibility to client drifts \cite{li2021model}. Thus we examine the trade-off between local epochs and communication rounds, as shown in Figure \ref{analysis:num_epochs}. We compare C2A with three baselines under the same number of model updates (local epochs $\times$ communication rounds). We observe that increasing the local epochs leads to worse performance due to the detrimental effect of client drift. Nevertheless, C2A clearly outperforms the other baselines in all settings. This further verifies the potency of C2A in mitigating the negative effects of the drift caused by excessive local updates, and shows that C2A can be efficiently trained with only a few rounds of communication.

\subsection{Communication Cost for Target Accuracy}

In FL scenarios, the communication between clients typically continues until the model attains a target accuracy or the allocated budgets are exhausted. As such, attaining the target accuracy with minimal communication rounds is crucial for reducing the total costs in practical FL. To analyze the baselines through the lens of such communication efficiency, we compare the number of required communications to reach the targeted performance for each baseline. The results are shown in Table \ref{exp:target_epochs}. Our proposed C2A consistently performs the best over the baselines on all target accuracy. Specifically, C2A reaches the targeted performance approximately two times faster than the vanilla adapter. These results show that C2A engages fewer communication costs with less requirement on the parameters and communication rounds.


\begin{table}[t]
\caption{The number of communication rounds required to achieve the desired performance. The relative speedup of each baseline is also compared to the vanilla Adapter \cite{adapter}.}
\adjustbox{width=\linewidth}{
\begin{tabular}{@{}lccc@{}}
\toprule
Methods   & Communication Rounds & & SpeedUp \\ \midrule
\multicolumn{4}{c}{\textit{Target accuracy = 70\%}} \\ \midrule
Adapter   & \phantom{11}13 \progressbar{50}{black} & \hspace{-3.8cm}\textcolor{red}{\vline height 0.28cm depth 3pt width 1.5pt}& $\times$1.00\phantom{1} \\ 
LoRA      & \phantom{11}18 \progressbar{73}{black} & \hspace{-3.8cm}\textcolor{red}{\vline height 0.28cm depth 3pt width 1.5pt}& $\times$0.72\phantom{1} \\ 
Compacter & \phantom{11}19 \progressbar{73}{black} & \hspace{-3.8cm}\textcolor{red}{\vline height 0.28cm depth 3pt width 1.5pt}& $\times$0.68\phantom{1} \\ 
Prompt-tuning    & \phantom{11}46 \progressbar{100}{black} & \hspace{-3.8cm}\textcolor{red}{\vline height 0.28cm depth 3pt width 1.5pt}& $\times$0.28\phantom{1} \\ 
BitFit    & \phantom{11}18  \progressbar{69}{black} & \hspace{-3.8cm}\textcolor{red}{\vline height 0.28cm depth 3pt width 1.5pt}& $\times$0.72\phantom{1} \\ 
AdaMix    & \phantom{11}12  \progressbar{45}{black} & \hspace{-3.8cm}\textcolor{red}{\vline height 0.28cm depth 3pt width 1.5pt}& $\times$1.10\phantom{1} \\ 


\cellcolor[gray]{.9}C2A (ours.) & \phantom{111}\textbf{7 }\progressbar{27}{black} & \hspace{-3.8cm}\textcolor{red}{\vline height 0.28cm depth 3pt width 1.5pt} & $\times$1.86\phantom{1} \\ \midrule

\multicolumn{4}{c}{\textit{Target accuracy = 80\%}} \\ \midrule
Adapter   & \phantom{11}33   \progressbar{50}{black} & \hspace{-3.8cm}\textcolor{red}{\vline height 0.28cm depth 3pt width 1.5pt}&$\times$1.00\phantom{1} \\ 
LoRA      & \phantom{11}44   \progressbar{66}{black} & \hspace{-3.8cm}\textcolor{red}{\vline height 0.28cm depth 3pt width 1.5pt}&$\times$0.75\phantom{1} \\ 
Compacter & \phantom{11}71   \progressbar{100}{black} & \hspace{-3.8cm}\textcolor{red}{\vline height 0.28cm depth 3pt width 1.5pt}&$\times$0.46\phantom{1} \\ 
Prompt-tuning  & 100$\uparrow$  \progressbar{100}{black} & \hspace{-3.8cm}\textcolor{red}{\vline height 0.28cm depth 3pt width 1.5pt}&$\times$0.33$\downarrow$ \\
BitFit    & \phantom{11}55   \progressbar{83}{black}& \hspace{-3.8cm}\textcolor{red}{\vline height 0.28cm depth 3pt width 1.5pt}& $\times$0.60\phantom{1} \\ 
AdaMix    & \phantom{11}50   \progressbar{76}{black}& \hspace{-3.8cm}\textcolor{red}{\vline height 0.28cm depth 3pt width 1.5pt}& $\times$0.66\phantom{1} \\ 
\cellcolor[gray]{.9}C2A (ours.)     & \phantom{11}\textbf{18} \progressbar{27}{black}& \hspace{-3.8cm}\textcolor{red}{\vline height 0.28cm depth 3pt width 1.5pt}& $\times$1.83\phantom{1} \\ \bottomrule
\end{tabular}
}
\label{exp:target_epochs}
\end{table}

\subsection{Scalability of C2A}
We evaluate whether C2A can be scaled to larger PLMs. To this end, we adopt all PEFT baselines to XLM-RoBERTa with 278M parameters. The results are summarized in Table \ref{exp:architecture}. We observe that our C2A still outperforms the baselines by a large margin. Specifically, our C2A achieves 3.1 points improvement compared with the adapter model. These results indicate that our approach can be well generalized to larger models. 

\subsection{Robustness to Client Drifts}



In order to showcase the robustness of C2A in non-IID scenarios, we employ CKA similarity to quantify the drift from the global model. Figure \ref{figure:motivation} shows that C2A is superior to other baselines in effectively reducing client drift. This justifies our hypothesis that creating tailored modules for each client is more effective in non-IID scenarios compared to a one-size-fits-all approach in training a single module for all clients.


\begin{table}[t]
\caption{Evaluation results of test accuracy (\%) with XLM-RoBERTa (278M) \cite{conneau2019unsupervised}. Best and second best results are highlighted in \textbf{boldface} and \underline{underlined}, respectively.}
\adjustbox{width=\linewidth}{
\begin{tabular}{@{}lcc@{}}
\toprule
Methods            & Params (\%) & Test Accuracy (\%) \\ \midrule
Full Fine-tuning   & 100\%         & 85.8                \\ \midrule
Adapter            & 0.217\%         & \underline{81.5}               \\
LoRA               & 0.106\%         & 80.7               \\
Prompt-tuning      & 0.008\%         & 65.8               \\
Compacter          & 0.021\%         & 77.7               \\
BitFit             & 0.037\%         & 79.7               \\
AdaMix             & 0.165\%         & 79.1               \\
\cellcolor[gray]{.9}C2A (ours.) & \cellcolor[gray]{.9}0.028\%         & \cellcolor[gray]{.9}\textbf{84.6}               \\ \bottomrule
\end{tabular}
}
\label{exp:architecture}
\end{table}


\section{Related Work}
\subsection{Parameter-efficient Fine-tuning}
Recent works on PEFT can be categorized into two lines of work: (1) tuning a subset of the existing parameters within the PLMs, including head fine-tuning \cite{freeze_last}, and bias tuning \cite{zaken2022bitfit}, (2) tuning with a small amount of additional trainable parameters, such as adapters \cite{adapter, compactor, adamix}, prefix-tuning \cite{prefix}, prompt-tuning \cite{prompt}, and low-rank adaption \cite{lora}.  Previous studies showed that PEFT achieves comparable performance compared to fine-tuning using only a small set of parameters. 
Given the advances brought by previous studies focused on centralized datasets, attention towards decentralized scenarios in FL remains under-explored. 
Yet, we discover that current PEFT approaches suffer from client drifts on non-IID setup, resulting in serious performance degradation in FL. Different from previous studies, we focus on improving the robustness of PEFT in decentralized scenarios by generating client-customized adapters.

\subsection{Federated Learning for NLP}
While much attention for FL has been focused on the field of computer vision, recent efforts have been done in applying FL to NLP tasks. For example, FedNLP \cite{fednlp} introduced benchmarks for evaluating FL methods and performed systematic analysis in the context of PLMs. \citet{pretrained-multi} examined FL in multilingual scenarios, where each client uses different languages. Similarly, several works attempted to extend the setting toward diverse tasks. For example, \citet{fedqa} adopted FL for question answering, and \citet{fedsentiment} proposed an aspect-based sentiment analysis method to enhance the performance under the restriction of data isolation. However, to the best of our knowledge, none of the prior works has been done on tackling the training complexity of FL on PLMs, which is directly related to the practicality.

\subsection{Hypernetworks in PEFT}
Prior studies have demonstrated that utilizing hypernetwork \cite{hypernetwork} is conducive to more efficient fine-tuning for PLMs in centralized scenarios. For instance, Hyperformer \cite{hyperadapter} and HyperPrompt \cite{hyperprompt} generated task-specific parameters by incorporating task-specific and layer-specific information on multi-task learning. Moreover, for multi-lingual learning, Hyper-X \cite{hyper-x} learned about the task and language-specific embeddings for generating adapters. While most previous works have been conducted for improving the efficiency of PEFT by utilizing the hypernetwork, they only focused on multi-task or multi-lingual situations. Instead, our work mitigates the client drifts issue of PEFT in federated scenarios by incorporating the data distributions of each client.

\section{Conclusion}
In this paper, we have observed significant performance degradation for typical PEFT approaches in decentralized scenarios. By carefully designed analysis, we have also shown that typical PEFT suffers from large client drifts, resulting in slow convergence and performance degradation. To address these issues, we have proposed C2A, a novel hypernetwork-based FL framework, which generates client-customized adapters by incorporating the data distribution of each client.
Our experimental results show that C2A achieves state-of-the-art results in various decentralized scenarios. Moreover, we have verified that C2A successfully mitigates the large client drift problem among local clients in FL scenarios.

\section{Limitations}
While we show that C2A successfully improves the effectiveness and efficiency of PEFT in FL, we have mainly focused on improving the effectiveness of the vanilla adapter. However, it is an open question whether our framework can improve other PEFT approaches, such as prompt tuning\cite{prompt}, and LoRA \cite{lora}. Although we didn't analyze whether our framework can generate parameters for alternative PEFT, one recent approach reveals that hypernetworks can generate parameters for various types of PEFT in multi-task learning \cite{hyperprompt, hyper-x}. Likewise, as C2A generates parameters with hypernetwork, we believe that C2A is highly expected to improve the performance of any alternative PEFT modules.

\section*{Ethics Statement}
This study covers work that utilizes PLMs, which have a wide variety of positive applications, such as the application to summarization or language understanding. At the same time, there are a number of ethical concerns with PLMs in general, including concerns regarding the generation of biased or discriminative text \cite{bias}, the leakage of private information from training data \cite{privacy}, and the environmental impact of training or tuning them \cite{enviroment}.

Our framework attempts to train PLMs with minimal changes made to their pre-existing parameters in FL scenarios. Our work is believed to bring some insights into the two ethical dimensions: privacy and environment. First, with respect to private information leakage, although our work has not addressed address the privacy issue in the pre-train process, our FL framework can mitigate the data privacy issues in the fine-tuning stages.
In addition, with respect to environmental impact, our work may obviate the need for full fine-tuning, which may also significantly reduce the cost in terms of memory or deployed servers.

\section*{Acknowledgment} 
This work was supported by the Basic Research Program
through the National Research Foundation of Korea
(NRF) grant funded by the Korea government (MSIT)
(2021R1A2C3010430) and Institute of Information \& Communications Technology Planning \& Evaluation (IITP) grant
funded by the Korea government (MSIT) (No. 2019-0-00079, Artificial Intelligence Graduate School Program (Korea University)). 


\appendix
\newpage

\begin{center}
\LARGE
\textbf{Supplementary Appendix}    
\end{center}


\section{Impact of Structure}
We analyze the effect with varied dimensions of the client embeddings and factorization in C2A. The detailed results are presented in Figure 5.

\paragraph{Effect of dimensions for client embeddings. } To investigate the effect of dimensions for client embeddings, we investigate the number of dimensions in C2A ranging from 1,4,8, and 32, during training. The results are shown in Figure 5(a). We observe that using a larger dimension of embeddings for adapters improves the training efficiency. Specifically, the model using eight dimensions shows the best performance. Thereby, we adopt a client embedding size of 8 in all our models.

\paragraph{Effect of dimensions for factorization. } Figure 5(b) represents the impact of latent dimensions for adapters in C2A. The dimension of factorization size 64 appears to be the best. Based on these results, we use an embedding size of 64 in all our models.

\begin{figure}[h]
\centering 
\includegraphics[width=1.0\linewidth]{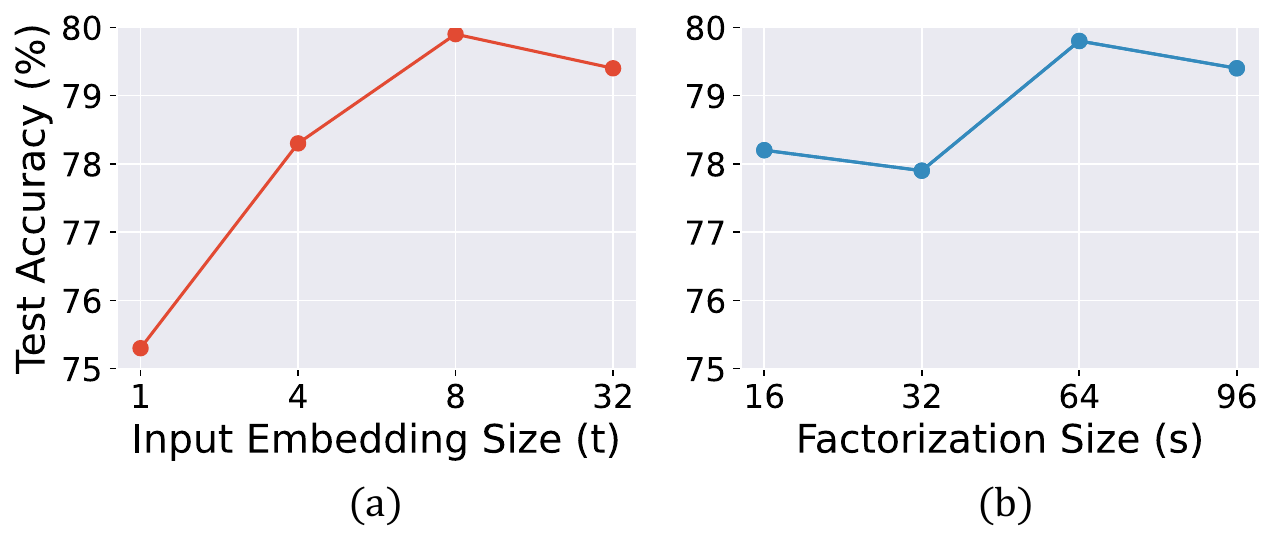}%
\caption{Evaluation results of test accuracy on NC dataset with the different number of dimensions for client embedding and factorization.}
\label{analysis:hyperparams}
\end{figure}

\section{Implementation details for C2A}
We implement C2A in Pytorch using four RTX 3090 GPUs for experiments with detailed hyper-parameter configurations as follows. We set the dimensionality of latent factors to $s = 64$ and client embeddings size of eight in all our models. Besides, for the low-rank dimension, we use a dimension of 16. 
We report the average results for all models of four random fine-tunings.

\end{document}